\documentclass{article}
\usepackage{spconf,amsmath,graphicx}
\usepackage{url}
\usepackage{xcolor}

\title{Image Coding for Machines with Edge Information Learning Using Segment Anything}
\name{Takahiro Shindo, Kein Yamada, Taiju Watanabe, Hiroshi Watanabe\thanks{The results of this research were obtained from the commissioned research (JPJ012368C05101) by National Institute of Information and Communications Technology (NICT), Japan.}}
\address{Graduate School of Fundamental Science and Engineering, Waseda University\\
Tokyo, Japan}
\begin{document}
%
\maketitle
\begin{abstract}
  Image Coding for Machines (ICM) is an image compression technique for image recognition.
  This technique is essential due to the growing demand for image recognition AI.
  In this paper, we propose a method for ICM that focuses on encoding and decoding only the edge information of object parts in an image, which we call SA-ICM.
  This is an Learned Image Compression (LIC) model trained using edge information created by Segment Anything.
  Our method can be used for image recognition models with various tasks.
  SA-ICM is also robust to changes in input data, making it effective for a variety of use cases.
  Additionally, our method provides benefits from a privacy point of view, as it removes human facial information on the encoder's side, thus protecting one's privacy.
  Furthermore, this LIC model training method can be used to train Neural Representations for Videos (NeRV), which is a video compression model.
  By training NeRV using edge information created by Segment Anything, it is possible to create a NeRV that is effective for image recognition (SA-NeRV).
  Experimental results confirm the advantages of SA-ICM, presenting the best performance in image compression for image recognition.
  We also show that SA-NeRV is superior to ordinary NeRV in video compression for machines.
  Code is available at {\color{magenta}\textit{\url{https://github.com/final-0/SA-ICM}}}.
\end{abstract}
\section{Introduction}
\label{sec:intro}
\begin{figure}[bt]
  \centerline{\includegraphics[width=1\columnwidth]{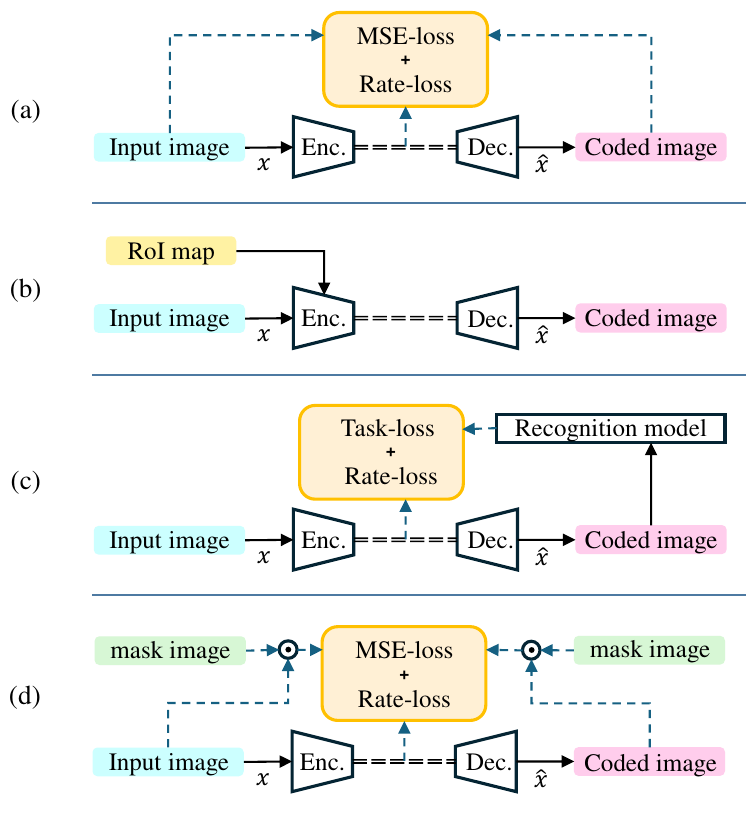}}
  \caption{Overview of image compression process. (a) : LIC model for human vision. (b) : ROI-based approach for ICM. (c) : Task-loss-based approach for ICM. (d) : Region-Learning-based approach for ICM.}
  \label{fig:process}
  \end{figure}

ICM is a technique for compressing images to reduce the bit rate without compromising image recognition accuracy.
Image compression technology is necessary for efficient transmission and storage of images, contributing to higher communication speeds and reduced device load.
Conventional image compression techniques are designed to encode and decode images with as little loss of visual image information as possible.
JPEG \cite{a1}, AVC/H.264 \cite{a2}, HEVC/H.265 \cite{a3} and VVC/H.266 \cite{a4} are standards that are constructed based on rule-based algorithms, designed to reduce the amount of information by primarily truncating the high-frequency components of an image.
This is based on the fact that the truncation of high-frequency components of an image has a small impact on image quality in human vision.
Apart from rule-based compression methods, there are also several image compression methods that use LIC models \cite{a5,a6}.
These models are trained to match the input and output images of the model, as shown in Fig.\ref{fig:process}(a).
The loss function is expressed by the following equation:
\begin{equation}
  \mathcal{L}_{h}=\mathcal{R}(y)+\lambda \cdot mse(x,\hat{x}).
  \end{equation}
  In (1), $y$ is the encoder output of the LIC model, $\mathcal{R}(y)$ is the bitrate of $y$ and is calculated using compressAI \cite{b1}. 
  $x$ represents the input image, and $\hat{x}$ represents the decoder output image.
  $mse$ represents the mean squared error function and $\lambda$ is a constant to control the rate.
These models also attempt to decode the pixel values of the input image, hence reconstructing images with good visual quality.
On the other hand, conventional methods are not compatible for ICM method.
This is because generally the amount of information in an image required for image recognition is less than that required for viewing \cite{b2}.
Therefore, research on ICM has been conducted, where JPEG and MPEG have begun standardization of image and video coding methods for machines.

There are three main approaches to ICM: Region of Interest (ROI)-based approach \cite{a11,a12,a28}, Task-loss (TL)-based approach \cite{a29,a30,a31}, and Region Learning (RL)-based approach \cite{c1}.
The ROI-based approach is a technique that uses an ROI-map to allocate more bits to a specific part of the image, as shown in Fig.\ref{fig:process}(b).
An image and its corresponding ROI-map are input to the encoder, and the image is compressed according to the guide of the map.
The problem with this approach is that the encoder must be equipped with an image recognition model to create the ROI-map.
The TL-based approach uses the output of the image recognition model as the loss function to train the LIC model, as shown in Fig.\ref{fig:process}(c).
The LIC model is trained to increase the image recognition accuracy of the decoded image.
Unfortunately, the LIC model is vulnerable to changes in image recognition models because it learns image compression methods for a particular image recognition model.
The RL-based approach is a method that allows the LIC model to learn only encoding/decoding methods for specific parts of the image, as shown in Fig.\ref{fig:process}(d).
The LIC model decodes object parts in the image cleanly and other parts roughly, enabling image compression for object detection and instance segmentation models.
Handmade mask images in the COCO dataset \cite{a33} are used to train this LIC model.
As for its concerns, image compression methods for semantic and panoptic segmentation models are not considered, since the decoded images do not preserve the background information in the images.

In this paper, we propose a new ICM model (SA-ICM) that solves the problems of the above approaches.
The proposed method is a type of RL-based approach, which does not require additional information such as ROI-map as encoder input and does not learn using task-loss.
On the other hand, unlike existing RL-based methods, it does not use mask images in the COCO dataset.
Instead, SA-ICM uses mask images generated by Segment Anything Model (SAM) \cite{c2}.
The edges of the segmentation map generated using SAM are used to train the LIC model.
This creates an LIC model that can decode only the main edge information.
The proposed method reduces more textures than existing RL-based approaches and while also removing human face textures.
It has good properties both in terms of image compression and privacy protection.
Furthermore, this learning method can be applied to the NeRV \cite{c3} learning method to create a video compression model for image recognition (SA-NeRV).
In experiments, we will investigate the image compression performance of SA-ICM for image recognition and compare it to other methods for ICM to demonstrate the effectiveness of this method.
For the image recognition models, we use an object detection model, an instance segmentation model, and a panoptic segmentation model to show the robustness to changes in the image recognition model.
By using COCO, VisDrone \cite{a37}, and Cityscapes \cite{a39} as the datasets, we show that our method can be used in various use cases.
We also compare the image recognition accuracy between SA-NeRV and ordinary NeRV decoded images to confirm the effectiveness of SA-NeRV.

\begin{figure*}[bt]
  \centerline{\includegraphics[width=2\columnwidth]{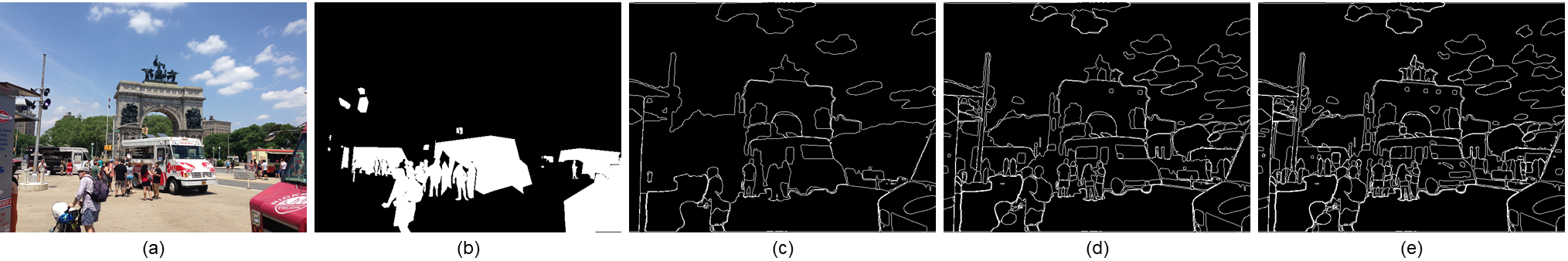}}
  \caption{Examples of the mask image. (a) : Original image. (b) : Mask image in COCO dataset. (c) : Mask image generated using SAM ($\alpha=0.98$). (d) : Mask image generated using SAM ($\alpha=0.93$). (e) : Mask image generated using SAM ($\alpha=0.48$).}
  \label{fig:mask}
  \end{figure*}
\begin{figure}[bt]
  \centerline{\includegraphics[width=1\columnwidth]{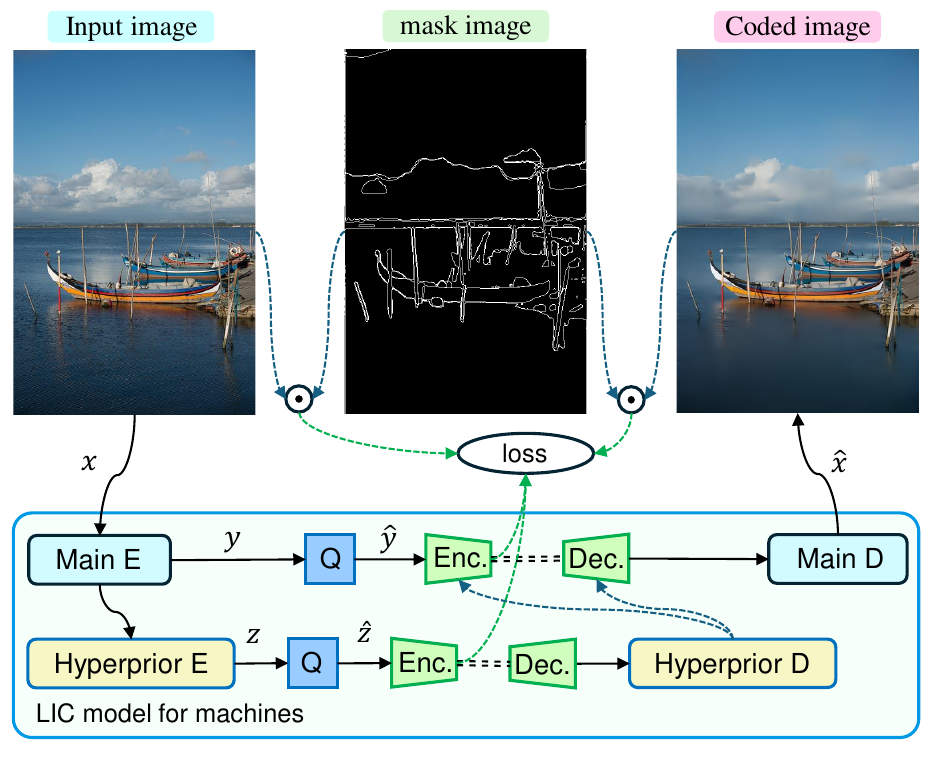}}
  \caption{The proposed training method of the LIC model.}
  \label{fig:train}
  \end{figure}

\section{Related Work}
\subsection{Image Coding for Machines (ICM)}

As opportunities for the use of image recognition technology increase, research on image compression for machines flourishes.
Many methods have been proposed, most of which can be categorized into one of the following three approaches, each with its drawbacks and advantages.
There are three main ICM approaches: the ROI-based approach, the TL-based approach, and the RL-based approach.

The ROI-based approach \cite{a11,a12,a28} uses an ROI-map to allocate more information to a specific part of the image, as shown in Fig.\ref{fig:process}(b).
This approach has the problem of placing a large load on the encoder's device because the ROI-map must be created on the encoder side.
Also, since more bits are allocated to the object part of the image, the decoded image is effective for object detection.
However, this approach is not necessarily effective for image recognition tasks that require a background.
On the other hand, the advantage of this approach lies in decoding images that are effective for both machine and human vision.
The study by B. Li \textit{et al}. \cite{a28} evaluates image compression performance in terms of object detection accuracy, instance segmentation accuracy, and image quality.

The TL-based approach \cite{a29,a30,a31} is an approach that attempts to optimize the LIC model using the output of the image recognition model, as shown in Fig,1(c).
Equation 1 plus task-loss, which is computed using the output of the image recognition model, is used as the loss function.
For example, to create an LIC model for YOLO \cite{a7}, a type of object detection model, the LIC model is trained using the object detection accuracy of YOLO in the decoded image.
The loss function for the LIC model in the TL-based approach is shown as:
\begin{equation}
  \mathcal{L}_{tl}=\mathcal{R}(y)+\lambda_{1} \cdot mse(x,\hat{x})+\lambda_{2} \cdot \mathcal{M}(\hat{x}).
\end{equation}
In (2), $\mathcal{R}$, $mse$, $y$, $x$, and $\hat{x}$ have the same meaning as those functions, variables, and constants in (1).
$\mathcal{M}(\hat{x})$ is the task-loss that can be computed by inputting the coded image into the image recognition model.
$\lambda_{1}$ and $\lambda_{2}$ are constants to control the rate.
The problem with this approach is that for a given image recognition model, a corresponding LIC model is required.
However, R. Feng \textit{et al}. \cite{a14} proposed an image compression method for various image recognition models using ResNet50 \cite{a32} feature-based learning method.

The RL-based approach \cite{c1} is the newest of these three approaches.
As shown in Fig.\ref{fig:process}(d), it is an ICM approach where the LIC model is trained to encode and decode only the texture of the object part in the image.
The loss function used to train the LIC model is the following:
\begin{equation}
  \mathcal{L}_{rl}=\mathcal{R}(y)+\lambda \cdot mse(x \odot m_{x},\hat{x} \odot m_{x}).
\end{equation}
In (3), $\mathcal{R}$, $mse$, $y$, $x$, $\hat{x}$, and $\lambda$ have the same meaning as those functions, variables, and constants in (1).
$m_{x}$ is the binary mask corresponding to $x$.
Handmade mask images in the COCO dataset are used as mask images.
This method has been shown to have good compression performance as an image compression method for object detection and instance segmentation models.
Conversely, the decoded images are not suitable for semantic and panoptic segmentation tasks because this LIC model does not learn how to encode and decode the background parts in the image.

\subsection{Neural Representations for Videos (NeRV)}
NeRV \cite{c3} is a technique for embedding video information in a neural network.
Unlike the conventional approach, which treats video as a collection of frame images, NeRV treats video as a neural network.
By inputting a frame index to the neural network corresponding to that video, the corresponding frame image is output.
The loss function used to train NeRV is expressed by the following equation:
\begin{equation}
  \mathcal{L}_{n}=\frac{1}{T}\sum_{t=1}^T\beta||x-\hat{x}||+(1-\beta)(1-ssim(x,\hat{x})).
\end{equation}
In (4), $x$, and $\hat{x}$ have the same meaning in (1). $T$ is the number of frames and $\beta$ is the constant to balance the weight for each loss componennt. 
By training NeRV so that the input image matches the output image, it is possible to decode images that are useful for human vision.
NeRV, which can embed video information, is applied as a video compression method.
By applying model pruning, model quantization, and weight encoding to the neural network in which the video is embedded, model compression is performed.
Since the neural network of NeRV is the video itself, compressing its model means to compress the video.
Experimental results have shown that this method has video compression performance comparable to HEVC.

\begin{figure*}[bt]
  \centerline{\includegraphics[width=2\columnwidth]{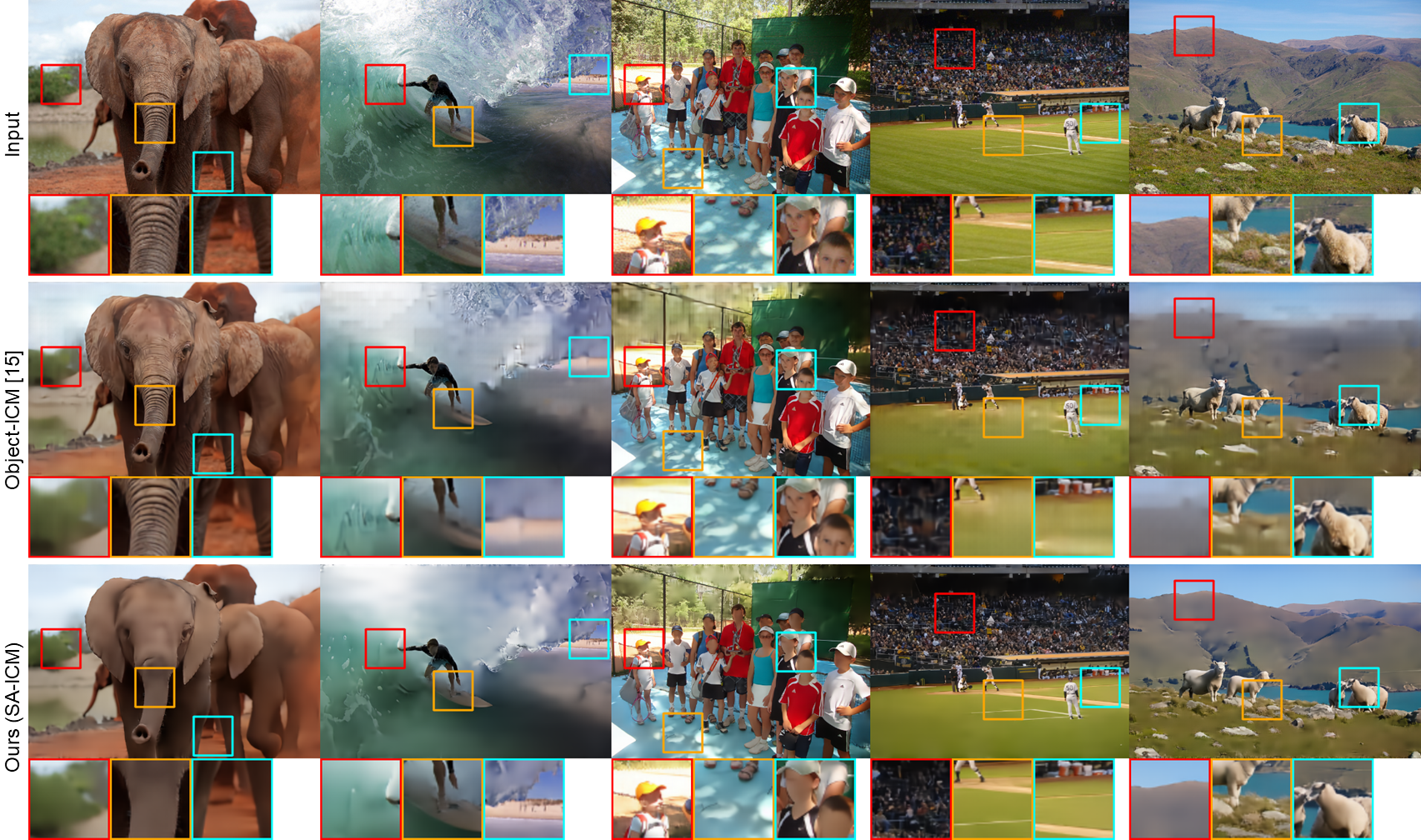}}
  \caption{Examples of coded images of the COCO2017 dataset. The top line is the input image, the middle line is the coded image by the conventional method of RL-based approach (Object-ICM)\cite{c1}, and the bottom line is the coded image by the proposed method (SA-ICM).}
  \label{fig:exp}
  \end{figure*}

\section{Proposed Method}
\subsection{SA-ICM}
We propose SA-ICM, a method to encode and decode only the edge information in an image by training the LIC model using a segment anything.
This method is a variant of the RL-based approach, which requires no additional information input and does not train the model using task-loss.
The original RL-based approach \cite{c1} used the image shown in Fig.\ref{fig:mask}(b) as the mask image and trained the LIC model only on the object regions in the image.
However, the decoded image using this approach is not suitable for image recognition tasks that require background information in the image because the background in the image is represented roughly.
In addition, unnecessary textures of object parts are decoded, meaning there are still rooms for improvement in compression performance.

In this paper, we train the LIC model using the images shown in Fig.\ref{fig:mask}(c)-(e) as mask images.
This mask image is obtained by inputting the segmentation map created from SAM into the Canny edge detector.
By changing the confidence value ($\alpha$) when estimating the segmentation map using SAM, different masks can be obtained, as shown in Fig.\ref{fig:mask}(c)-(e).
The smaller the $\alpha$, the more object masks SAM outputs, hence more edges are detected, as shown in Fig.\ref{fig:mask}(e).
As shown in Fig.\ref{fig:train}, the LIC model trained with these masks learns to encode and decode only the edge information in the image.
The loss function used to train the LIC model is expressed as follows:
\begin{gather}
  \mathcal{L}_{p}=\mathcal{R}(y)+\lambda \cdot mse(x \odot sam_{x}(\alpha),\hat{x} \odot sam_{x}(\alpha)).
\end{gather}
In (5), $\mathcal{R}$, $mse$, $y$, $x$, $\hat{x}$, and $\lambda$ have the same meaning as those functions, variables, and constants in (1).
$sam_{x}(\alpha)$ is the mask image corresponding to $x$ created using SAM.
These mask images are only used during training of the LIC model and are not used during testing.
This learning method creates an LIC model capable of removing object texture while not completely removing background information.

\subsection{SA-NeRV}
The learning method of SA-ICM is applied to NeRV to improve the image recognition accuracy in NeRV decoded images.
The original NeRV \cite{c3} is a technique to embed video information necessary for human vision into a neural network using Eq.(4) as a loss function.
In this paper, we propose a technique to embed video, especially the edges of the video, into a neural network (SA-NeRV).
The loss function used to train SA-NeRV is as follows:
\begin{gather}
  \mathcal{L}_{sa-n}=\mathcal{L}_{n}+\frac{1}{T}\sum_{t=1}^T\beta||x\odot sam_{x}(\alpha)-\hat{x}\odot sam_{x}(\alpha)|| \notag \\
  +(1-\beta)(1-ssim(x\odot sam_{x}(\alpha),\hat{x}\odot sam_{x}(\alpha))).
\end{gather}
In (6), $x$, $\hat{x}$, $T$ and $\beta$ have the same meaning as those variables and constants in (4).
$\mathcal{L}_{n}$ is the loss component in NeRV training, as shown in Eq. (4).
By learning NeRV with $\mathcal{L}_{sa-n}$ as the loss function, the position and shape of objects in the video are efficiently embedded in the neural network.
This method can be used to decode images that are useful for image recognition models.

\begin{figure*}[bt]
  \begin{minipage}[b]{0.48\linewidth}
      \centering{\includegraphics[width=1\columnwidth]{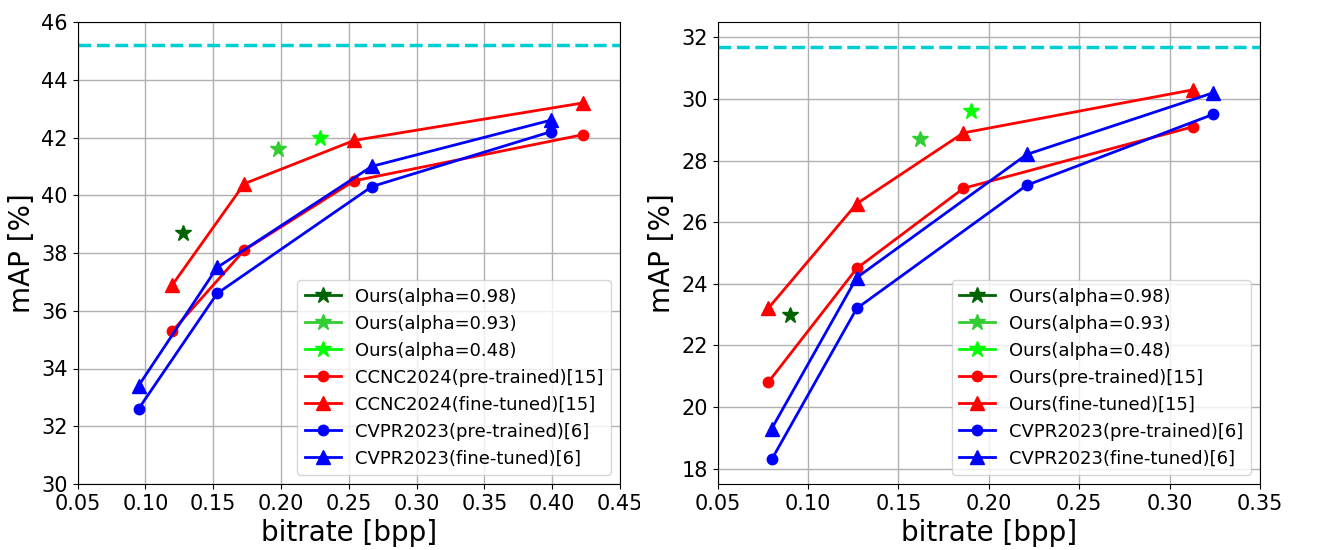}}
      \caption{Compression performance in object detection accuracy of YOLOv5. The left figure shows compression performance for COCO, and the right figure shows the same for VisDrone.}
      \label{fig:coco}
    \end{minipage}
  \begin{minipage}[b]{0.48\linewidth}
      \centering{\includegraphics[width=1\columnwidth]{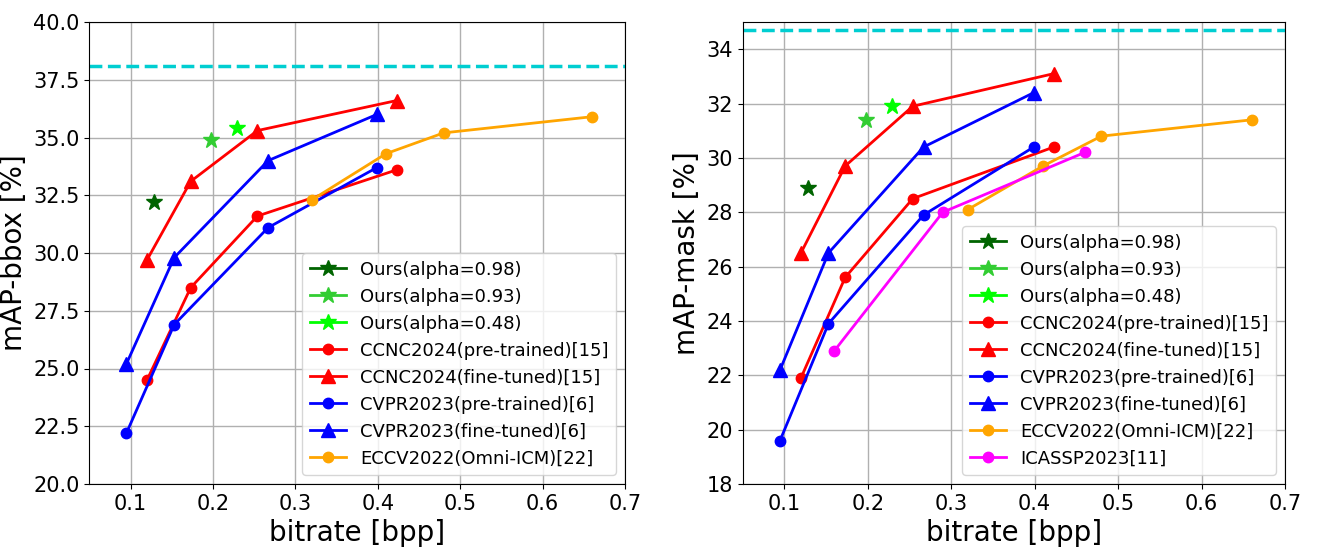}}
      \caption{Compression performance in image recognition accuracy of Mask-RCNN. The left and right figures show the compression performance in detection accuracy and instance segmentation accuracy, respectively.}
      \label{fig:rcnn}
    \end{minipage}
  \end{figure*}
\begin{figure*}[bt]
  \begin{minipage}[b]{0.48\linewidth}
      \centering{\includegraphics[width=1\columnwidth]{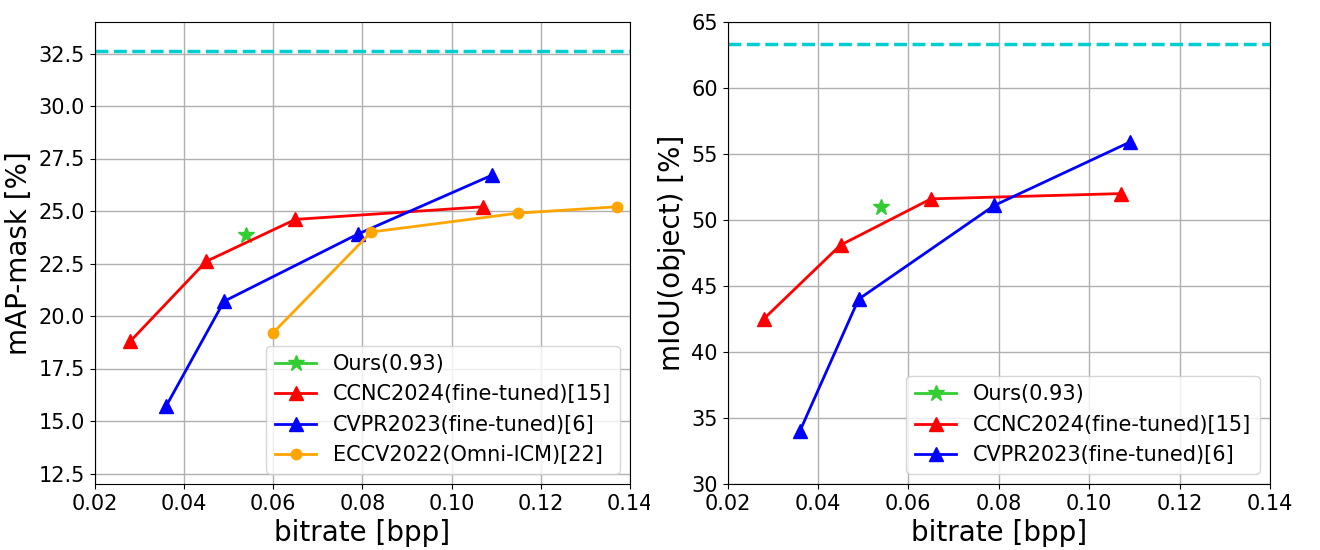}}
      \caption{Compression performance in instance segmentation accuracy of Panoptic-deeplab.}
      \label{fig:city}
      \end{minipage}
  \begin{minipage}[b]{0.48\linewidth}
      \centering{\includegraphics[width=1\columnwidth]{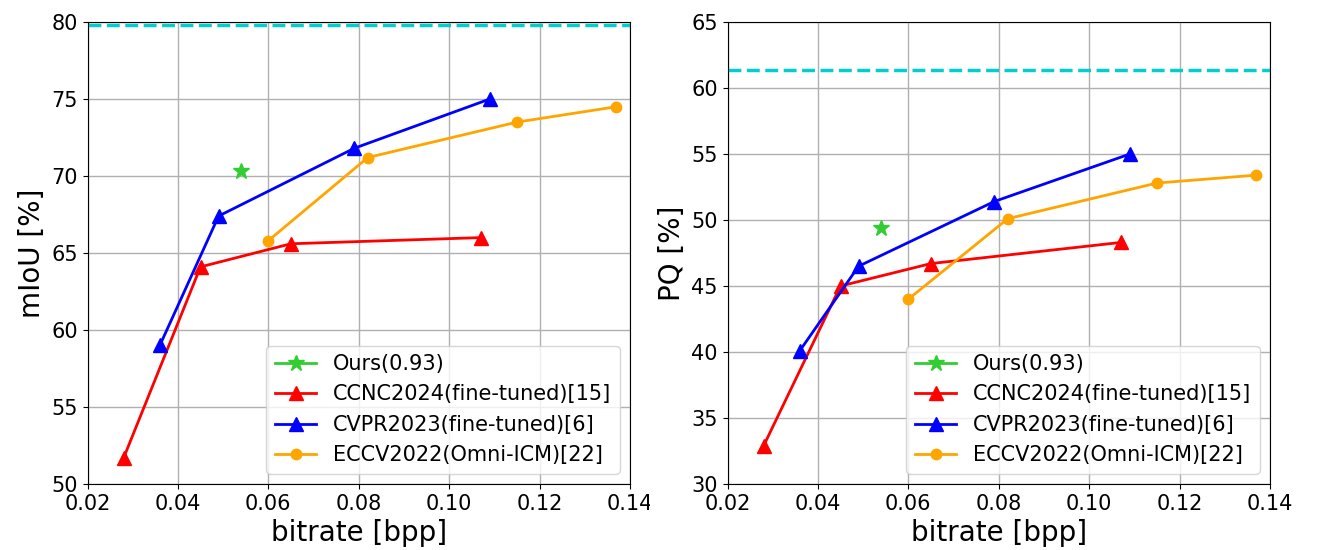}}
      \caption{Compression performance in panoptic segmentation accuracy of Panoptic-deeplab.}
      \label{fig:city1}
      \end{minipage}
  \end{figure*}

\section{Experiments}
\subsection{Experimental Methods for Evaluating SA-ICM}
To confirm the effectiveness of SA-ICM, we measure its image compression performance.
First, a mask image corresponding to an image in the COCO-train dataset is created using SAM.
The confidence values used for mask creation are [0.98,0.93,0.48].
These mask images are used to train the LIC model.
We use the model proposed by J. Liu \textit{et al}. \cite{a6} for the LIC model and Eq.(5) for the loss function.
Although the rate can be controlled by changing $\lambda$, we set the value to 0.05 in this experiment.
The decoded image with these models is shown in Fig.\ref{fig:exp}.
It can be seen that the decoded image loses its texture but the information of object shape is retained.
Also, the information on the human face is lost during compression, which is good for privacy protection.
Next, we measure the image compression performance of the trained LIC models.
YOLOv5 \cite{a34}, Mask-RCNN \cite{a35}, and Panoptic-deeplab \cite{a36} are used as image recognition models.
Mask-RCNN can simultaneously perform instance segmentation and object detection, while Panoptic-deeplab can perform panoptic segmentation and instance segmentation at the same time.
The object detection accuracy when using YOLOv5 is measured by the COCO dataset and the VisDrone dataset.
This YOLOv5 is fine-tuned by data composed of compressed training datasets, obtained from trained LIC model.
The training data for each dataset is compressed using the LIC model, and YOLOv5 is fine-tuned with those data.
The image recognition accuracy when using Mask-RCNN and Panoptic-deeplab is measured using the COCO and Cityscapes datasets, respectively.

\subsection{SA-ICM Evaluation Experimental Results}
Comparisons of the proposed method with other methods for ICM are shown in Fig.\ref{fig:coco}-\ref{fig:city1}.
In all these figures, the light blue dotted line represents the image recognition accuracy in uncompressed images, and the green star-shaped points indicate the image compression performance of the proposed method.
Originally, many points are calculated by varying the value of $\lambda$ in Eq.(5), but in this experiment, the compression performance at multiple points is calculated by changing $\alpha$ instead of $\lambda$.
Fig.\ref{fig:coco} shows the relationship between object detection accuracy and bit rate for the COCO and VisDrone datasets.
The model used for object detection is YOLOv5.
In both figures, mAP50:95 is used as the index of object detection accuracy.
Fig.\ref{fig:rcnn} shows the image recognition accuracy using Mask-RCNN on the COCO dataset.
The left figure shows the relationship between object detection accuracy and bitrate.
The right figure represents the relationship between instance segmentation accuracy and bitrate.
It can be seen that the proposed method has better compression performance than conventional RL-based methods in the object detection and instance segmentation tasks.
Fig.\ref{fig:city} and Fig.\ref{fig:city1} show the relationship between image recognition accuracy and bit rate using Panoptic-deeplab.
Unlike the original RL-based ICM method, SA-ICM has good image compression performance for panoptic segmentation tasks that require background information in the image.
These results indicate that SA-ICM is an effective image compression method for image recognition models of various tasks, robust to changes in use cases.

\begin{figure}[bt]
  \centerline{\includegraphics[width=0.95\columnwidth]{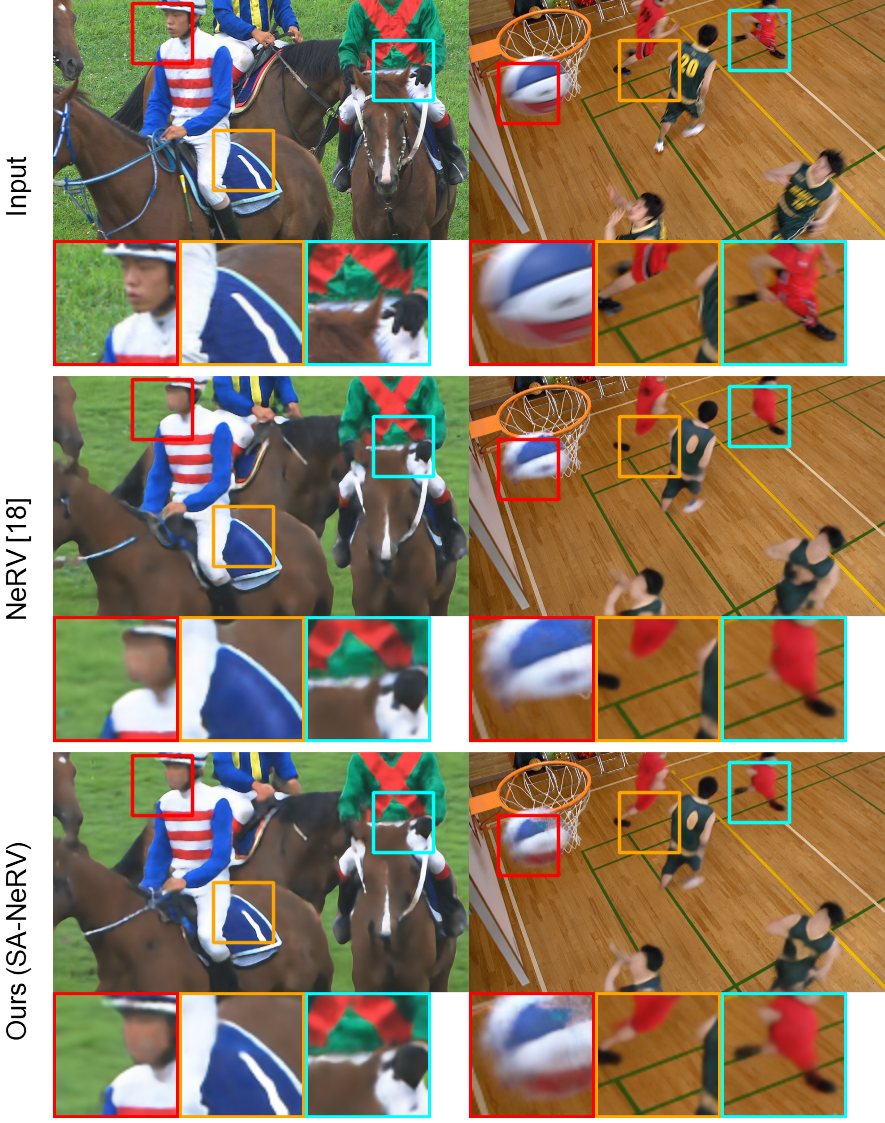}}
  \caption{The top line is the input video frame, the middle line is the decoded frame using NeRV \cite{c3}, and the bottom line is the decoded frame using SA-NeRV.}
  \label{fig:nerv}
  \end{figure}
\subsection{Experimental Methods for Evaluating SA-NeRV}
To evaluate the efficiency of SA-NeRV, we compare the image recognition accuracy of the original NeRV decode images and SA-NeRV decoded images.
We use the SFU-HW-Objects-v1 \cite{a37} dataset, consisting 18 sequences and their corresponding annotations for object detection.
These sequences are classified into five classes (A to E) according to the image size and the characteristics.
This dataset is also applied for the Common Test Condition in MPEG's VCM standardization activities.
In this experiment, we work with class C and class D sequences from this data set.
After embedding these videos in NeRV and SA-NeRV, the videos are decoded.
We use the pre-trained YOLOv7 \cite{a38} to measure the object detection accuracy in the decoded videos.

\subsection{SA-NeRV Evaluation Experimental Results}
The object detection accuracies in the decoded images are shown in Table \ref{tab:nerv}.
It can be seen that for most sequences, the detection accuracy in the SA-NeRV decoded image is higher than that in the NeRV decoded image.
An example of a decoded image is shown in Fig.\ref{fig:nerv}.
The decoded image of SA-NeRV has a more correct decoded object shape than the decoded image of the original NeRV.
From the above, it can be said that the decoded image of SA-NeRV is more appropriate than the decoded image of the existing method in terms of image recognition.

\begin{table}[t]
  \centering
  \caption{Object detection accuracy (mAP [\%]) of NeRV and SA-NeRV decoded video in each sequence.} \label{tab:nerv}
  \small
  \begin{tabular*}{8cm}{@{\extracolsep{\fill}}p{3cm}|p{2cm}|p{2cm}}
    \hline
    \qquad sequence name & \quad NeRV \cite{c3}  & \quad SA-NeRV\\

    \hline
    \hline
    \quad BQMall          & \qquad 28.03 & \qquad \textbf{28.24}\\
    \quad BasketballDrill & \qquad 34.26 & \qquad \textbf{34.93}\\
    \quad PartyScene      & \qquad 34.34 & \qquad \textbf{34.61}\\
    \quad RaceHorsesC     & \qquad 80.99 & \qquad \textbf{81.77}\\
    \quad BQSquare        & \qquad 27.84 & \qquad \textbf{29.80}\\
    \quad BasketballPass  & \qquad 23.29 & \qquad \textbf{24.88}\\
    \quad BlowingBubbles  & \qquad 41.83 & \qquad \textbf{48.84}\\
    \quad RaceHorsesD     & \qquad \textbf{89.12} & \qquad 88.98\\
    \hline
  \end{tabular*}
\end{table}

\section{Conclusion}
In this paper, we propose SA-ICM and SA-NeRV. 
Using edge information learning, we constructe an LIC model that encodes and decodes object shapes in images. 
Compared to conventional methods, our LIC model reveals superior image compression performance. 
Other than the benefit from a privacy point of view, our method is also flexible to change in use cases. 
Furthermore, we confirm that the image recognition accuracy of the NeRV-decoded image can be improved by employing our training method.

\end{document}